\pdfoutput=1

\documentclass[11pt]{article}

\usepackage{acl}

\usepackage{times}
\usepackage{latexsym}
\usepackage{graphicx}
\usepackage[T1]{fontenc}

\usepackage[utf8]{inputenc}

\usepackage{microtype}

\title{Stability of Syntactic Dialect Classification Over Space and Time}

{\author{Jonathan Dunn$^1$ \and Sidney Wong$^2$ \\
         Department of Linguistics \\
         University of Canterbury \\
         $^1$New Zealand Institute for Language, Brain and Behaviour \\
         $^2$Geospatial Research Institute Toi Hangarau \\
         \texttt{jonathan.dunn@canterbury.ac.nz} \\ \texttt{sidney.wong@pg.canterbury.ac.nz}
         }}

\begin{document}
\maketitle
\begin{abstract}
This paper analyses the degree to which dialect classifiers based on syntactic representations remain stable over space and time. While previous work has shown that the combination of grammar induction and geospatial text classification produces robust dialect models, we do not know what influence both changing grammars and changing populations have on dialect models. This paper constructs a test set for 12 dialects of English that spans three years at monthly intervals with a fixed spatial distribution across 1,120 cities. Syntactic representations are formulated within the usage-based Construction Grammar paradigm (CxG). The decay rate of classification performance for each dialect over time allows us to identify regions undergoing syntactic change. And the distribution of classification accuracy within dialect regions allows us to identify the degree to which the grammar of a dialect is internally heterogeneous. The main contribution of this paper is to show that a rigorous evaluation of dialect classification models can be used to find both variation over space and change over time.
\end{abstract}

\section{Geographic Variation Over Time}

This paper experiments with the stability of dialect classification models over space and time in order to determine the degree to which they capture language variation and change. The assumption in previous work has been that a geo-referenced corpus \cite{df15,Cook2017,Dunn2020} captures the linguistic behaviour of specific populations. This paper experiments with the spatial and temporal stability of dialect models by systematically constructing monthly test sets spanning a three-year period. This allows us to evaluate the continuing effectiveness of dialect models over time, an important criteria for determining their validity. Because different locations represent different populations, we use spatial sampling to construct test sets which represent different local populations within each country. This allows us to determine the degree to which a dialect like New Zealand English adequately represents the varied populations within New Zealand.

Dialect classification is the task of predicting the location of origin for the individual who produced a given sample \cite{10.3389/frai.2019.00015, chakravarthi-etal-2021-findings-vardial, gaman-etal-2020-report}. Thus, dialect classification, by focusing on the latent properties of geo-referenced samples, differs from geo-location \cite{Rahimi2017} which focuses on predicting the location of the sample itself and from geo-characterization \cite{Adams2018} which focuses on predicting attributes of the location. While all three tasks rely on geographic information, dialect classification is unique in modelling variations in the linguistic system. Beyond this, dialect classification is part of ensuring that \textsc{nlp} represents the world's population, including non-standard and non-western populations.

The temporal evaluation (Section 6) shows that most dialects share the same performance decay rate. This indicates a general effect of model decay rather than cases of change over time within individual dialects. The spatial evaluation, however, shows that prediction accuracy for all dialects is spatially-conditioned within countries (Section 7). This indicates that, while dialect models capture proto-typical populations within each country, they do not equally describe all local populations.

The experiments in this paper use construction grammar (CxG: \citealt{g06a, l08, Croft2013}) to represent syntactic structure for the purpose of observing dialectal variation. CxG is a \textit{usage-based} approach to syntax, a bottom-up theory of language in which frequent exposure is hypothesized to lead to the emergence of grammatical units \citep{Hopper1987,Bybee2006}. The use of syntactic representations for dialect classification ensures that the model does not rely on extraneous information like place names or local topics of interest. From this perspective, a \textsc{grammar} is a set of constructions that together represent the structure of a language. A \textsc{dialect model} is a matrix of spatial weights in which the number of rows corresponds to the number of constructions in the grammar and the number of columns corresponds to the number of dialects. These weights, learned using a Linear \textsc{svm}, support dialect classification and also represent spatial variation in the grammar.

In order to undertake a spatio-temporal evaluation, we collect a balanced corpus of tweets to represent 12 varieties of English around the world. The basic experimental paradigm is to train models on a fixed period (July through December 2018) and then test those models at monthly intervals from 2019 to 2021. Each monthly test set maintains the same geographic distribution as the training data, so that fluctuations in performance are not caused by changes in the locations represented.

After considering related work on dialect classification and other geographic models (Section 2), we consider the corpora used in these experiments (Section 3). We then present the syntactic representations used (Section 4) and the basic experimental methods (Section 5). The performance of dialect models over time is presented in Section 6 and the performance over space in Section 7. The main contribution of this paper is to show that the performance of dialect classification models remains stable over time but that there is significant spatial variation in performance within dialect areas.

\section{Related Work}

Early work showed that part-of-speech trigrams are able to distinguish between some regional dialects \cite{s07}, a method that continues to appear in recent work \cite{kd18}. Similar methods have been used for authorship analysis \cite{hf07} and for characterizing immigrant populations \cite{nerbonne-wiersma-2006-measure}. In other contexts, non-syntactic features can out-perform syntactic features for modelling dialects \cite{k}, so that many approaches to distinguishing between dialects are similar to language identification models \cite{Ali2018}.

More recent work has modelled geographic syntactic variation by combining grammar induction with geospatial text classification \cite{d18b,10.3389/frai.2019.00015,Dunn2019a}. The use of grammar induction to learn a syntactic feature space mitigates the fact that most grammars represent standard varieties \cite{Jorgensen2015}, thus poorly representing many dialects around the world. In this paradigm, the learned grammar provides a feature space (c.f., Section 4) and the frequency of grammatical constructions in each sample is used to model dialects: a bag-of-constructions approach to text classification.

\begin{table*}[t]
\centering
\begin{tabular}{|lllll|}
\hline
\textbf{Circle} & \textbf{Region} & \textbf{Country} & \textbf{N. Cities} & \textbf{N. Words} \\
\hline
Inner-Circle & Oceania & Australia & 98 & 3.9 mil \\
Inner-Circle & Oceania & New Zealand & 99 & 2.0 mil \\
Inner-Circle & North American & Canada & 95 & 4.9 mil \\
Inner-Circle & North American & United States & 86 & 4.5 mil \\
Inner-Circle & European & Ireland & 100 & 3.6 mil \\
Inner-Circle & European & United Kingdom & 89 & 5.5 mil \\
\hline
\textbf{Total Inner-Circle} & \textbf{3} & \textbf{6} & \textbf{567} & \textbf{24.4 mil} \\
\hline
\hline
Outer-Circle & African & Ghana & 69 & 1.1 mil \\
Outer-Circle & African & Kenya & 98 & 1.8 mil \\
Outer-Circle & South Asian & India & 96 & 2.5 mil \\
Outer-Circle & South Asian & Pakistan & 100 & 1.0 mil \\
Outer-Circle & Southeast Asian & Malaysia & 99 & 0.8 mil \\
Outer-Circle & Southeast Asian & Philippines & 91 & 1.1 mil \\
\hline
\textbf{Total Outer-Circle} & \textbf{3} & \textbf{6} & \textbf{553} & \textbf{8.37 mil} \\ 
\hline
  \end{tabular}
  \caption{Inventory of Regions, Countries, and Cities for Data Collection (One Month)}
  \label{tab:1}
\end{table*}

Most work on geographic variation is focused on lexical variation \cite{eosx10} and change \cite{eosx14}. Recent work has shown a close correspondence between lexical variation in tweets and lexical variation in a dialect survey \cite{10.3389/frai.2019.00011}. This work is important for showing that digital usage mirrors face-to-face usage. Other work has shown that geographic variation can be taken into account during language identification to ensure the inclusion of non-standard varieties \cite{Jurgens2017}. Models of lexical variation have generally failed to account for polysemy, so that competition between senses is not captured \cite{Zenner2012}, but more recent work has been able to account for polysemy in this context \cite{Lucy2021}. 

A related line of work uses language data to model non-linguistic properties of populations and places. For example, the problem of geo-location is to predict the location of a user given properties of a document \cite{Wing2014,Alex2016,Rahimi2017}. This task differs from dialect classification in that named entities and topic features can provide significant information. A related task is to model the characteristics of a particular place rather than the population of that place \cite{Adams2015,Adams2018,Hovy2018,Villegas2020}. While there is a close connection between a place and its population, this line of work remains focused on characterizing non-linguistic attributes.

This paper makes two main contributions: \textit{First}, it experiments with geographic syntactic variation over time and within dialect regions, significantly expanding our understanding of geographic variation in syntax. \textit{Second}, from a more practical perspective, this paper evaluates the degree to which geographic models remain robust over space and time, an evaluation not previously available.

\section{Geographic Language Data}

This paper draws on social media data from the \textit{Corpus of Global Language Use (\textsc{cglu})}, using geo-referenced tweets that are identified for language using the idNet package \cite{Dunn2020}. The collection method for social media in the \textsc{cglu} involves geographic searches from co-ordinates of individual cities. Here we sample from 1,120 cities representing 12 countries and six regions, as shown in Table \ref{tab:1}. This table shows the amount of data by place by month. The total data set contains six months for training and 36 months for testing. The data set as a whole is visualized at \href{https://www.earthlings.io}{earthLings.io}.

This corpus is designed to provide a balanced representation of different varieties of English over time. The colonial history of English has led to a distinction within the World Englishes paradigm \cite{k90} between \textit{inner-circle} varieties that represent the first diaspora (e.g., Canada) and \textit{outer-circle} varieties that represent the second diaspora (e.g., India). We include six dialects/varieties each from the inner-circle and outer-circle groups.

Within each group we include three regions, each with two country-level varieties. As shown in Table \ref{tab:1}, the inner-circle group contains three regions: Oceania (Australia and New Zealand), North America (Canada and the US), and Europe (the UK and Ireland). The collection of data from these countries is distributed across 567 cities, where each city represents a 50km radius from the city center. For each month, we sample 24.4 million words representing these inner-circle varieties. 

The outer-circle group also contains three regions: Africa (Ghana and Kenya), South Asia (India and Pakistan), and Southeast Asia (Malaysia and the Philippines). The collection of data from these countries is distributed across 553 cities, with a comparable sample of 8.37 million words for each month across the training and testing periods.


To maintain a comparable geographic distribution over time, we maintain the same number of samples from each city. This means, for example, that the relative influence of Brisbane and Perth in Australia remain constant over time. A \textit{sample} for the purposes of this paper is an aggregation of individual tweets from the same place and time until the sample reaches 500 words. These larger samples provide more syntactic information for each dialect than do individual tweets. While previous work has used samples of 1,000 words \citep{10.3389/frai.2019.00015}, here we use smaller samples in order to increase the capacity for error analysis. As with many tasks, there is a trade-off between the higher accuracy provided by larger sample sizes and the flexibility provided by smaller sample sizes.

The distribution of samples across cities is taken from the training period (2018). Thus, the density of data by location across time is fixed to represent the density during the training period. This allows us to control for changes in the collection: for example, if Wellington began to produce more data in 2021, this change in distribution within New Zealand would appear to be syntactic variation while actually reflecting a change in the means of observation. Data collection spans from 07-2018 until 12-2021, a period of 42 months. The training period is 2018 and the testing period is 2019 through 2021. The geographic distribution across countries, as shown in Table \ref{tab:1}, is held constant across this period, controlling for other sources of variation that might impact dialect models.

\section{Syntactic Representations}

This section details the main ideas of construction grammar (CxG), including both (i) the grammar induction algorithm used to learn syntactic representations here and (ii) examples of constructions used in the dialect models. The basic approach here is, first, to use grammar induction to learn a grammar and, second, to use the frequency of the constructions in that grammar to undertake geospatial text classification \cite{10.3389/frai.2019.00015, Dunn2019a}. 

CxG can be distinguished from other approaches to syntax given its three core ideas: First, CxG posits a continuum between the lexicon and the grammar rather than a strict separation (for example, into a vocabulary and a set of phrase structure rules). This \textsc{constructicon} contains both lexical items and traditional syntactic structures. For example, a grammar-and-lexicon approach would analyze (a) below as an intransitive sentence by labelling the verb \textit{laugh} as intransitive. The problem is that verb valency is quite fluid, as shown in (b) and (c). The CxG analysis of this fluidity is that (a) represents an \textsc{intransitive} construction into which \textit{laugh} is merged and (b)/(c) represent a \textsc{caused-motion} construction into which \textit{laugh} is merged. Thus, the fluidity of the argument structure here is explained by an underlying construction, itself meaningful, which interacts with specific lexical items. (Note that the grammar used for modelling dialects does not contain any individual lexical items as constructions).

~

(a) Peter \textit{\underline{laughed}}.

(b) The audience \textit{\underline{laughed}} Peter off the stage.

(c) His marriage \textit{\underline{laughed}} Peter into rehab.

(d) Peter \textit{\underline{laughed}} all the way to the bank.

~

A second main idea in CxG is that syntactic structure varies in its level of abstractness, with some representations being quite item-specific. The constructicon is an inheritance hierarchy in which fully-productive constructions like the \textsc{caused-motion} construction in (b)/(c) have item-specific children like the idiom in (d). Essentially, (d) is a non-compositional and idiomatic version of the construction in (b)/(c) with some of the slots constrained to  require a fixed phrase.

~

\noindent (e) [\textsc{syn:np} -- \textsc{syn:vp}]

\noindent (f) [\textsc{syn:np} -- \textsc{syn:vp} -- \textsc{sem}:\textit{object}  -- \textsc{sem}:\textit{loc}]

\noindent (g) [\textsc{syn:np} -- \textsc{syn:vp} -- \textsc{lex}:\small\textit{all the way to the bank}\normalsize]

A third main idea in CxG is that constructions are constraint-based representations in which slot-fillers are drawn from lexical, syntactic, and semantic categories. Each unit in a construction is a \textit{slot}, separated by dashes in (e)/(f)/(g) above. Each slot is defined using a \textit{slot-constraint}. For example, the \textsc{intransitive} construction in (e) can be represented using only syntactic constraints. In contrast, the \textsc{caused-motion} construction in (f) has two semantic constraints; these are labelled for purposes of exposition as \textit{object} and \textit{location}. The construction in (g) is item-specific and idiomatic, so that it can only be described using lexical constraints. The point, then, is that different levels of abstraction are captured in CxG using different types of slot-constraints.

This paper draws on previous approaches to the unsupervised learning of constructions \citep{d17,d18}. The first challenge is to build the inventory of lexical, syntactic, and semantic constraints that constructions are built on. Here we use the most frequent 100k words across the entire corpus of tweets as the lexicon. The syntactic constraints are drawn from the Universal Part-of-Speech tagset \citep{pdm12} as implemented by the Ripple-Down-Rules tagger \citep{nn}. The semantic constraints are drawn from fastText embeddings \citep{Grave2019} clustered into discrete semantic domains using k-means. A complete inventory of these semantic domains is provided in \href{https://jdunn.name/2022/09/12/stability-of-syntactic-dialect-classification-over-space-and-time/}{the supplementary material}; this approach ignores polysemy in lexical items when defining semantic constraints, using a single representation for each word-form.

From the perspective of varying levels of abstractness, syntactic constraints are the most general because they are divided into the smallest inventory of labels (only 14). Lexical constraints are the least general, with a lexicon of 100k words. And semantic constraints are in the middle, with an inventory of 1,000 domains. This parameter choice (i.e., using 1,000 semantic domains) results from the desired granularity in domains, falling between the very general syntactic constraints and the very specific lexical constraints. Thus, constructions are a sequence of slots, each of which is defined by a slot-constraint. Each type of slot-constraint (lexical, semantic, and syntactic) differ in their level of abstractness. For instance, lexically-defined constructions are more idiomatic and item-specific than syntactically-defined constructions.

This work relies on a loss function based on Minimum Description Length \citep{Goldsmith2006,GrunwaldP.andRissanen2007} and a construction parser with a beam-search strategy \citep{Dunn2019z} that operates on top of a psychologically-plausible association measure, the $\Delta P$ \citep{Ellis2007}. The contribution of this paper is to analyze syntactic variation across space over time using previous work on computational CxG; thus, we do not provide a fuller description of the framework here. Previous work has shown that these grammars converge onto stable representations as the amount of training data is increased \citep{Dunn2021}, that grammars of individuals are significantly different than grammars of groups of individuals \citep{Nini2021}, and that transformer-based language models can be fine-tuned using constructional information \citep{tayyar-madabushi-etal-2020-cxgbert}.

The grammar used in these experiments is learned from the training period (2018) but includes a wider pool of 18 English-speaking countries in order to provide a global grammar of English. This larger training corpus for grammar induction contains 478 million words. The fastText embeddings are trained on this same extended corpus, but covering the entire period in order to increase the amount of data available for training; this larger corpus contains 4.2 billion words. This results in a single grammar that contains 6,119 individual constructions, some of which are shown in (h) through (n) below. Dialect models are learned by parsing each sample using this grammar, counting the frequency of each construction in each sample, and using the resulting feature space for dialect classification. The complete grammar, along with examples from the training data for each construction, is available in \href{https://jdunn.name/2022/09/12/stability-of-syntactic-dialect-classification-over-space-and-time/}{the supplementary material}.

\begin{figure*}[t]
\centering
\includegraphics[width = 460pt]{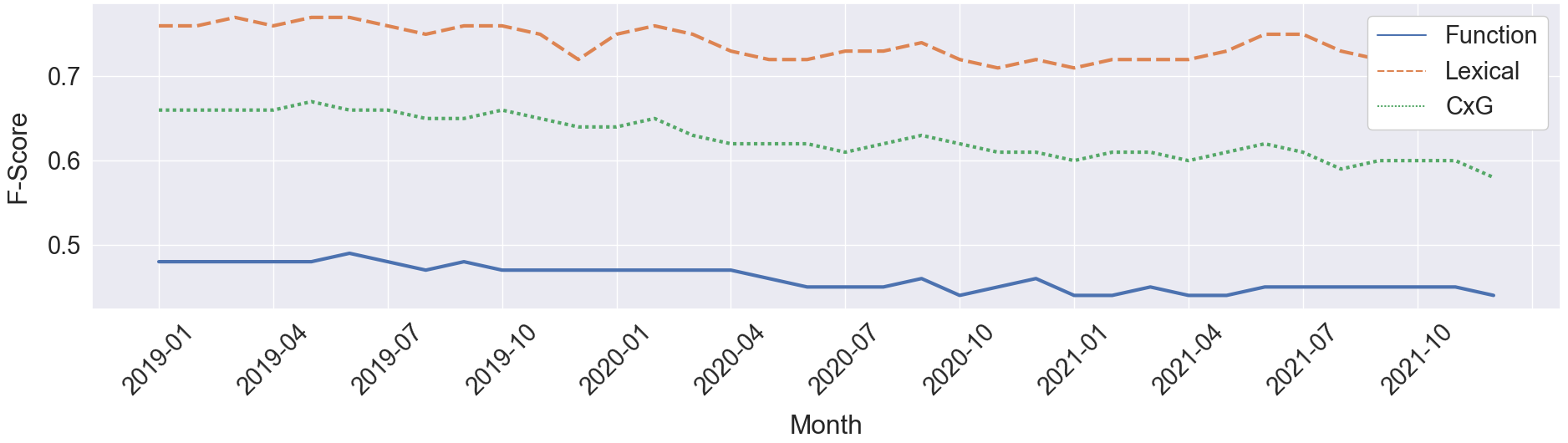}
\caption{F-Score Against Baselines Over Time, All Varieties}
\label{fig1}
\end{figure*}

The following examples illustrate the nature of constructions; both constructions like (h) and examples like (h1) are drawn from the grammar used in the experiments. Each slot in (h) is separated by dashes and each slot-constraint is defined using lexical (\textsc{lex}), syntactic (\textsc{syn}), or semantic (\textsc{sem}) categories. Lexical constraints are words given in italics; syntactic constraints are drawn from part-of-speech tags; and semantic constraints are formulated using numbers that refer to clustered embeddings, such as <443> in (k). For dialect classification, each construction (h) provides a feature and the frequency of that construction (h1 through h3) provides a sample-specific quantification.

\noindent (h) [\textsc{lex}:\textit{it} -- \textsc{syn}:\textsc{aux} -- \textsc{syn}:\textsc{v}]

(h1) `it is set'

(h2) `it was shut'

(h3) `it can go'

~

The first example, in (h), shows a simple clause with an expletive \textit{it} as subject and a variable auxiliary verb. The example in (i) is a lexically-constrained noun phrase with \textit{ability} as the head of an infinitival verb. A further lexically-constrained noun phrase in (j) shows the importance of a tweet-specific grammar: \textit{ur} replaces the more traditional \textit{your} as the pronoun. 

~

\noindent (i) [\textsc{lex}:\textit{ability} -- \textsc{lex}:\textit{to} -- \textsc{syn}:\textsc{v}]

(i1) `ability to focus'

(i2) `ability to live'

(i3) `ability to wait'

~

\noindent (j) [\textsc{lex}:\textit{ur} -- \textsc{syn}:\textsc{adj} -- \textsc{syn}:\textsc{n}]

(j1) `ur new journey'

(j2) `ur own money'

(j3) `ur mad tunes'

~

The adposition phrase in (k) contains a semantic constraint on the complement noun, in this case a type of location. As an example of how constructions themselves can be meaningful, (l) shows a copula construction with an ending conjunction. But the construction as a whole marks a caveat on the evaluation that is expressed by the copula.

~

\noindent (k) [\textsc{syn}:\textsc{adp} -- \textsc{syn}:\textsc{n} -- <443>]

(k1) `along airport road'

(k2) `in union station'

(k3) `into police station'

~

\noindent (l) [\textsc{syn}:\textsc{n} -- \textsc{lex}:\textit{was} -- \textsc{syn}:\textsc{adj} -- \textsc{syn}:\textsc{cc}]

(l1) `bike was awesome but'

(l2) `birthday was great and'

(l3) `movie was better but'

~

The more complicated verb phrase in (m) contains a main verb, \textit{myself} as a direct object, and an infinitival verb. This implicitly constrains the main verb to verbs of thinking like \textit{compare} and \textit{tell}, showing that implicit semantic constraints arise from interactions between slots. Finally, the complex noun phrase in (n) reflects a specific template of \textsc{np + adp}. In this way, constructions capture grammatical units of varying size and abstractness.

\noindent (m) [\textsc{syn}:\textsc{v} -- \textsc{lex}:\textit{myself} -- \textsc{lex}:\textit{to} -- \textsc{syn}:\textsc{v}]

(m1) `allowing myself to hope'

(m2) `forcing myself to sleep'

(m3) `tell myself to stop'

~

\noindent (n) [\textsc{lex}:\textit{the}--\textsc{syn}:\textsc{n} -- \textsc{lex}:\textit{of} -- \textsc{syn}:\textsc{det} -- \textsc{syn}:\textsc{n}]

(n1) `the happiness of another person'

(n2) `the owner of the station'

(n3) `the masters of the game'

~

\begin{figure*}[t]
\centering
\includegraphics[width = 460pt]{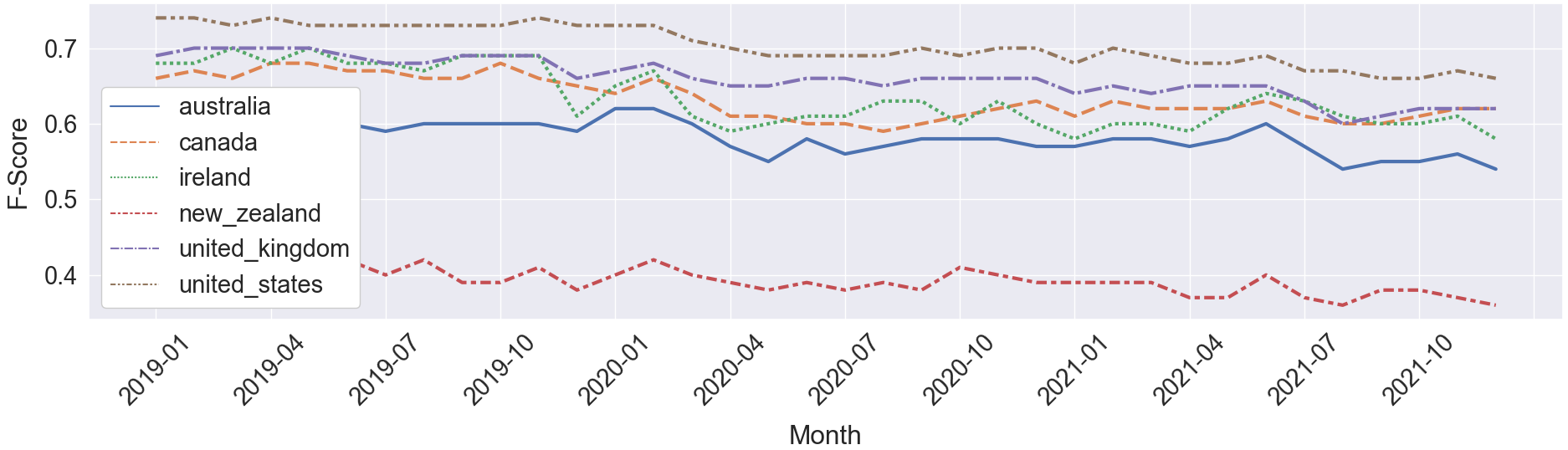}
\caption{F-Score by Country Over Time for CxG Model, Inner-Circle Varieties}
\label{fig2}
\end{figure*}

This section has presented CxG as a paradigm for usage-based syntax and reviewed previous work on computational CxG. An unsupervised construction grammar is learned from the training period, providing an adaptable feature space that contains structures from many different dialects. As the discussed examples show, these learned constructions provide a rich syntactic feature space for modelling geographic variation in production over time. 

\section{Dialect Models}

The task of dialect classification or identification is to predict the location of origin for the author of a sample given some set of linguistic features. The classification here predicts country-level dialects like New Zealand English or Australian English. From the perspective of linguistics, dialect classification allows us to study variation in a high-dimensional space: variation across an entire grammar \cite{10.3389/frai.2019.00015} rather than variation in individual and independent features \cite{10.3389/frai.2019.00011}. From the perspective of \textsc{nlp}, dialect classification is part of the general problem of ensuring that language technology represents the world's population rather than privileged sub-sets of the world's population \cite{dunn-adams-2020-geographically}.

Because part of the goal is to model spatio-temporal variation in the grammar, a dialect model takes the form of a matrix in which each feature (a construction in the grammar) is a row and each dialect (a country-level label) is a column. This matrix represents the degree to which a given part of the grammar is subject to geographic variation. Taken row-wise, this matrix provides a measure of whether a particular construction varies across space. And, taken column-wise, this matrix provides a description of each dialect that, for example, can be compared with every other dialect to determine which are the most similar. As discussed below, dialect models are implemented as Linear \textsc{svm}s that are trained using a bag-of-constructions approach in which the parser counts how many times each construction occurs in each sample.

Using the data from 2018 for training, we compare three models: \textit{First}, a syntactic model based on the frequencies of the constructional features described above. \textit{Second}, a baseline model that uses the frequency of function words like \textit{of} or \textit{was}, a common baseline for problems in authorship analysis \citep{Grieve2007,Stamatatos2009,Argamon2018} when content words need to be avoided. \textit{Third}, for the purpose of comparison, we include a unigram lexical model with \textsc{tf-idf} weighting and function words removed so that it contains no syntactic information. Each of these models are implemented as a Linear \textsc{svm}. Within this task, \textsc{svm}s remain competitive, as shown by recent shared tasks on Romanian dialect identification \citep{gaman-etal-2020-report} and on identifying similar Uralic languages \citep{chakravarthi-etal-2021-findings-vardial}.  In each case, we use a development set to determine parameters.

In each case, we train three models: \textsc{inner} contains only inner-circle varieties like American English; \textsc{outer} contains only outer-circle varieties like Indian English; and \textsc{all} contains all 12 varieties. These are trained on the data from 2018 and tested on data from 2019, 2020, and 2021. The reason for maintaining separate models in some conditions is that inner-circle varieties have significantly more training and testing data available, which could lead to higher performance as an artifact. Thus, for example, the inner-circle condition contains only training and testing data from the six countries listed as \textit{inner-circle} in Table \ref{tab:1}.

As an initial analysis, the f-scores of each of these three models over time is shown in Figure \ref{fig1}, with the y-axis indicating the weighted f-score and the x-axis indicating time. All three classifiers are well above the majority baseline. The lowest performing is the function word model, a weak approximation for syntactic variation. The highest performing is the lexical model. This hierarchy remains stable across the three year testing period.

\begin{table}[h]
\centering
\begin{tabular}{|llll|}
\hline
\textbf{AU} & \textbf{CA} & \textbf{IE} & \textbf{NZ} \\
\hline
australia & canada & ireland & nz \\
australian & canadian & irish & zealand \\
mate & ontario & dublin & auckland \\
melbourne & trump & cork & jacinda \\
sydney & toronto & limerick & te \\
abc & vancouver & galway & kiwi \\
brisbane & trudeau & lads & liked \\
labor & km & hurling & lincoln \\
nsw & kpa & county & hamilton \\
turnbull & alberta & final & kph \\
\hline
  \end{tabular}
  \caption{Top Lexical Features By Country}
  \label{tab:2}
\end{table}

Given the results in Figure \ref{fig1}, could we use the lexical model to examine dialects? The issue, as in previous work, is that the information contained in this model does not represent linguistic variation. Table \ref{tab:2} shows the top lexical items for four inner-circle countries: Australia, Canada, Ireland, and New Zealand. Most of these terms are place-names (like \textit{australia}), place-specific named-entities (like \textit{abc}), or people associated with these countries (like \textit{jacinda}). Only a few terms would qualify as dialectal variants, for example \textit{mates} vs \textit{lads}. As a representation of latent linguistic variation, the lexical model is not relevant; we thus focus on the syntactic models in the remaining analysis.


\section{Syntactic Variation Over Time}

We begin the analysis by looking at the weighted average f-score by model for the beginning of the test period (2019-01) and the end (2021-12), as shown in Table \ref{tab:3}. This represents the impact of time on the overall accuracy. First, we see that outer-circle models have better performance. The most likely reason for this is that outer-circle varieties are more distinct from one another, in part because these varieties exist in more linguistically-diverse settings. For example, the US is less linguistically diverse than India in digital settings \citep{dunn-etal-2020-measuring}. Although outer-circle varieties have a higher average f-score, they also have a greater change in f-score. This indicates more variability over time.

\begin{table}[h]
\centering
\begin{tabular}{|lcc|}
\hline
\textbf{~} & \textbf{Function} & \textbf{Grammar} \\
\hline
Inner-Only, 2019-01 & 0.44 & 0.66 \\
Inner-Only, 2021-12 & 0.40 & 0.59 \\
\hline
\textit{Inner-Only Decline} & \textit{0.04} & \textit{0.07} \\
\hline
\hline
Outer-Only, 2019-01 & 0.75 & 0.83 \\
Outer-Only, 2021-12 & 0.66 & 0.75 \\
\hline
\textit{Outer-Only Decline} & \textit{0.09} & \textit{0.08} \\
\hline
\hline
All Dialects, 2019-01 & 0.48 & 0.66 \\
All Dialects, 2021-12 & 0.44 & 0.58 \\
\hline
\textit{All Dialects Decline} & \textit{0.04} & \textit{0.08}\\
\hline
  \end{tabular}
  \caption{Change in Performance Over Time by Model}
  \label{tab:3}
\end{table}

Second, we notice in Table \ref{tab:3} that the relative performance of function words and the CxG model remain similar across the testing period. The full grammar model always out-performs the function word baseline. We use a regression analysis to model the decay rate for each dialect in the CxG models, examining the amount of change in precision and recall over time (c.f., Figure \ref{fig2}). The basic idea here is that a consistent decay rate indicates model error while a faster rate of decay for individual dialects indicates change in those dialects themselves. Among inner-circle varieties, only \textsc{nz} has a significant difference from the others, for recall but not for precision. A decline in precision would mean that samples from other dialects have become more similar to \textsc{nz}; this does not happen. The observed decline in recall means that samples from \textsc{nz} have become more similar to other dialects. This indicates that there has been a significant change in \textsc{nz} but not in other dialects. No outer-circle varieties have a different decay rate, so that only \textsc{nz} shows this type of change.

\begin{table}[h]
\centering
\begin{tabular}{|l|cccccc|}
\hline
~ & \textbf{\textsc{au}} & \textbf{\textsc{ca}} & \textbf{\textsc{ie}} & \textbf{\textsc{nz}} & \textbf{\textsc{uk}} & \textbf{\textsc{us}} \\
\hline
\textbf{\textsc{au}} & 	0	& 0	& 0	& 0	& 0	& 0 \\
\textbf{\textsc{ca}} & 	0	& 0	& 0	& 0	& 0	& 0 \\
\textbf{\textsc{ie}} &   0	& 0	& 0	& 0	& 0	& -.04 \\
\textbf{\textsc{nz}} & .16	& 0	& 0	& 0	& .28	& 0 \\
\textbf{\textsc{uk}} & 	0	& 0	& 0	& 0	& 0	& 0 \\
\textbf{\textsc{us}} & 	0	& 0	& 0	& 0	& 0	& 0 \\
\hline
\hline
~ & \textbf{\textsc{gh}} & \textbf{\textsc{in}} & \textbf{\textsc{ke}} & \textbf{\textsc{my}} & \textbf{\textsc{pk}} & \textbf{\textsc{ph}} \\
\hline
\textbf{\textsc{gh}} & 	0	& -1.01	& 0	& 0	& -.28	& -.40 \\
\textbf{\textsc{in}} & 	0	& 0	& 0	& -.91	& 0	& 0 \\
\textbf{\textsc{ke}} &   0	& 0	& 0	& 0	& 0	& 0 \\
\textbf{\textsc{my}} &   0	& 0	& 0	& 0	& 0	& 0 \\
\textbf{\textsc{pk}} & 	-.20	& 0	& 0	& 0	& 0	& .13 \\
\textbf{\textsc{ph}} & 	0	& 0	& 0	& 0	& 0	& 0 \\
\hline
  \end{tabular}
  \caption{Changing Relationships Between Dialects \\
  Using a \textsc{vecm} Analysis of False Positive Errors}
  \label{tab:4}
\end{table}

The decay rate represents the overall trend for a given dialect but it does not take into account the specific errors made. The confusion matrix for each dialect provides a monthly representation of the distribution of false positive errors. For example, in the CxG model that includes all dialects, Canadian English has 1,488 false positives as American English in the first test period, but only 48 with India and 7 with Pakistan. This distribution of false positive errors over time provides a more detailed view of the classifier's performance. Because the classification model itself does not change after training, changes in the distribution of errors reflect changes that have arisen in a given dialect after training.

The question here is whether the relationship between dialects (geographic variation) changes over time. We model this using a Vector Error Correction Model (\textsc{vecm}: \citealt{time_series}). This model checks for relationships between multiple time series, which in this case reflect changing error patterns between dialects. The data represents a non-stationary time series because the number of errors in all dialects increases over time (i.e., there is a decline in performance as shown in Table \ref{tab:3}). To partially control for the increase in errors over time, we examine the relative frequency of false positives by country by month. The \textsc{vecm} model allows us to determine if there is a significant long-term trend in the distribution of errors from a given dialect, robust to short-term variations.

We examine the significant changes by country for the inner-circle and outer-circle models with CxG features in Table \ref{tab:4}. Only significant changes are shown; negative values indicate that samples for the row have become more frequently mistaken for the column. Thus, for the inner-circle varieties, \textsc{nz} becomes more similar over time to Australia and the UK. This means that, in addition to lower classification performance, \textsc{nz} is also subject to the most change in the way it is situated among other dialects. Outer-circle varieties on the whole are subject to more change in error distribution over time than inner-circle varieties. The analysis of decay rates also shows that \textsc{nz} was subject to change over time; the difference is that this analysis takes into account the distribution of errors rather than viewing the error rate as a black box. The outer-circle varieties have a changing error distribution, but not a changing error rate.

\section{Syntactic Variation Within Countries}

While previous work has viewed a dialect area as a homogeneous entity, here we have sampled from approximately 100 points for each country and maintained a consistent sample over time. To what degree is the performance of dialect classifiers driven by geographic trends within a country? If a country like Australia has a single dominant grammar, then the performance of the syntax-based classifier should be relatively consistent within that country. To test this hypothesis, we look at the average accuracy over time for samples collected from each point within a country.

\begin{table}[h]
\centering
\begin{tabular}{|l|c|ccc|}
\hline
~ & \textbf{Moran's \textit{I}} & \textbf{Mean Acc.} & \textbf{Min} & \textbf{Max} \\
\hline
\textbf{\textsc{au}} & 0.30 & 61\% & 18\% & 83\% \\
\textbf{\textsc{ca}} & 0.54 & 65\% & 07\% & 100\% \\
\textbf{\textsc{ie}} & 0.17 & 58\% & 35\% & 89\% \\
\textbf{\textsc{nz}} & 0.20 & 36\% & 08\% & 62\% \\
\textbf{\textsc{uk}} & 0.22 & 73\% & 41\% & 82\% \\
\textbf{\textsc{us}} & 0.18 & 79\% & 53\% & 97\% \\
\hline
\hline
~ & \textbf{Moran's \textit{I}} & \textbf{Mean Acc.} & \textbf{Min} & \textbf{Max} \\
\hline
\textbf{\textsc{gh}} & 0.30 & 86\% & 42\% & 94\% \\
\textbf{\textsc{in}} & 0.38 & 84\% & 27\% & 95\% \\
\textbf{\textsc{ke}} & 0.24 & 89\% & 62\% & 97\% \\
\textbf{\textsc{my}} & 0.70 & 79\% & 50\% & 95\% \\
\textbf{\textsc{pk}} & 0.42 & 70\% & 15\% & 87\% \\
\textbf{\textsc{ph}} & 0.20 & 77\% & 37\% & 88\% \\
\hline
  \end{tabular}
  \caption{Geographic Variation in Performance}
  \label{tab:5}
\end{table}

This is shown in Table \ref{tab:5} with a global Moran's \textit{I} used as a measure of spatial autocorrelation within a country \cite{Anselin1988}. A common method in geospatial statistics, Moran's \textit{I} measures the correlation in a single variable (here, prediction accuracy for dialect classification) across different locations. This measure has values closer to 1 when the variable is highly spatially organized and closer to 0 when there is no spatial organization. Given that there are different numbers of samples from each location, it is possible that a generic Moran's \textit{I} would view sparse locations as outliers; thus, we use the Empirical Bayes rate adjustment to control for the level of precision in each location as well \cite{Xia1998, Anselin2006}. 

The table also shows the mean accuracy across cities and the min and max accuracy. These results show that there is an effect for location: the dialect models work well in some places and not so well in others. The Moran's \textit{I} determines whether this variation in performance is spatially structured. Because different locations represent different populations, these are measures of how well the dialect models work for the entire population of a country. Full maps and spatial results are available in \href{https://jdunn.name/2022/09/12/stability-of-syntactic-dialect-classification-over-space-and-time/}{the supplementary material}.

\begin{figure}[t]
\centering
\includegraphics[width = 230pt]{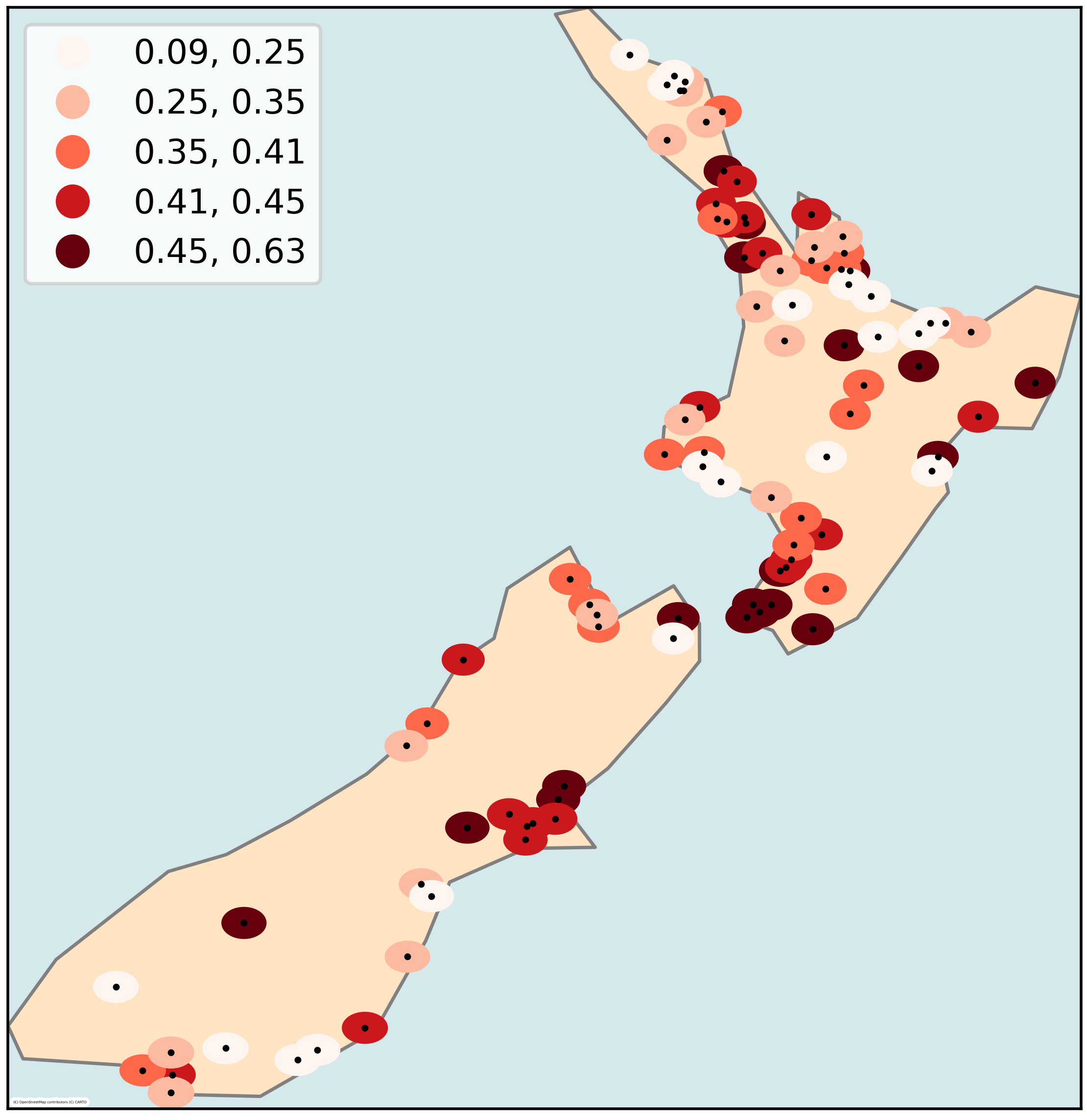}
\caption{Map of Average City-Level Accuracy, \textsc{nz}}
\label{fig3}
\end{figure}

All countries have a significant spatial pattern to their accuracy distribution. Within inner-circle countries, Canada has the highest deviation, with a wide range in accuracy and a significant spatial structure to that variation. The \textsc{us} and \textsc{uk} have the highest accuracy, while \textsc{nz} performs much worse than other dialects, perhaps because of the change over time discussed above. To explore this further, we visualize the internal variation for \textsc{nz}, the inner-circle dialect with the lowest performance and the most change over time, in Figure \ref{fig3}. Each collection point is a dot and the shading in the surrounding radius represents the accuracy for that collection area. Darker colors represent higher accuracy. The main cities (Auckland, Wellington, Christchurch) have the most consistent performance. But areas with known distinct linguistic landscapes like Northland (far north) and Southland (far south) have much lower accuracy. More rural areas around the country have consistently lower accuracy as well. The main point in this spatial error analysis is that, because different locations represent different populations, the observed variations in accuracy show that these dialect models do not equally represent all populations within the country.

\section{Conclusions}

This paper has shown that syntax-based dialect classifiers can reveal both spatial and temporal patterns in linguistic variation. We find that the models remain robust over time, with a fixed decay rate, with the exception of change observed in \textsc{nz}. This means that, while classification performance does decline, the rate of decline is predictable and evenly distributed. Within dialect regions, however, there is a significant spatial effect on performance. This evaluation is important for establishing an understanding of how dialect models and other geographic models function in the face of on-going linguistic change and population change over space and time. Here, even the best dialect models do not equally represent all speakers of a dialect.

\bibliography{anthology,custom}
\bibliographystyle{acl_natbib}

\end{document}